\documentclass{article}

\usepackage{microtype}
\usepackage{graphicx}
\usepackage{subcaption}
\usepackage{booktabs} 

\usepackage{textcomp}
\usepackage{xcolor}
\usepackage{amsmath}

\usepackage{hyperref}

\usepackage[preprint]{icml2026}

\usepackage{amsmath}
\usepackage{amssymb}
\usepackage{mathtools}
\usepackage{amsthm}

\usepackage{multirow}
\usepackage{algorithm}
\makeatletter
\@ifpackageloaded{algorithmic}{%

}{}
\makeatother
\usepackage{algorithmicx}
\usepackage{algpseudocode}
\usepackage{xargs}
\usepackage{xurl}
\usepackage{makecell}
\usepackage{threeparttable}

\usepackage{enumitem}

\usepackage[capitalize,noabbrev]{cleveref}

\theoremstyle{plain}

\theoremstyle{definition}

\theoremstyle{remark}

\usepackage[textsize=tiny]{todonotes}

\newcommandx{\HaoNote}[2][1=]
{\todo[inline,linecolor=black,backgroundcolor=green!25,bordercolor=green,#1]{#2 ---HL}}

\icmltitlerunning{Ask the Expert: Collaborative Inference for Vision Transformers with Near-Edge Accelerators}

\newif\ifsubmission
\submissiontrue 

\ifsubmission
\usepackage{xpatch}
\xapptocmd\normalsize{%
 \abovedisplayskip=0pt plus 1pt minus 1pt
 \abovedisplayshortskip=0pt plus 1pt
 \belowdisplayskip=0pt plus 1pt minus 1pt
 \belowdisplayshortskip=1pt plus 1pt minus 1pt
}{}{}
\newcommand{\smartparagraph}[1]{%
  \par\vspace{0pt plus 0.5pt}
  \noindent\textbf{#1}
}

\newenvironment{smartlist}{%
    \begin{list}{\textbf{$\bullet$}}{%
        \setlength{\topsep}{-\parskip}\addtolength{\topsep}{0pt plus 0.5pt}
        \setlength{\partopsep}{0pt plus 0.5pt}%
        \setlength{\itemsep}{0pt plus 0.5pt}%
        \setlength{\parsep}{0pt plus 0.5pt}%
        \setlength{\leftmargin}{1.7em}
        \setlength{\labelwidth}{1.2em}
        \setlength{\labelsep}{0.5em}
        \setlength{\listparindent}{0pt}%
    }%
}{%
    \end{list}%
}

\newlist{smartenum}{enumerate}{1}
\setlist[smartenum]{
    topsep=0pt,
    partopsep=0pt,
    itemsep=0pt,
    parsep=0pt,
    leftmargin=1.7em,
    labelwidth=1.2em,
    labelsep=0.5em,
    label=\textbf{\arabic*.}
}

\else
\fi

\begin{document}

\twocolumn[
  \icmltitle{Ask the Expert: Collaborative Inference for Vision Transformers with Near-Edge Accelerators}

  \icmlsetsymbol{equal}{*}

  \begin{icmlauthorlist}
    \icmlauthor{Hao Liu}{yyy}
    \icmlauthor{Suhaib A. Fahmy}{yyy}
  \end{icmlauthorlist}

  \icmlaffiliation{yyy}{CEMSE Division, King Abdullah University of Science and Technology, Thuwal 23955, Saudi Arabia}

  \icmlcorrespondingauthor{Suhaib A. Fahmy}{suhaib.fahmy@kaust.edu.sa}

  \icmlkeywords{Machine Learning, ICML}

  \vskip 0.3in
]

\printAffiliationsAndNotice{}  

\begin{abstract}
Deploying Vision Transformers on edge devices is challenging due to their high computational complexity, while full offloading to cloud resources presents significant latency overheads. 
We propose a novel collaborative inference framework, which orchestrates a lightweight generalist ViT on an edge device and multiple medium-sized expert ViTs on a near-edge accelerator. 
A novel routing mechanism uses the edge model's Top-$\mathit{k}$ predictions to dynamically select the most relevant expert for samples with low confidence. 
We further design a progressive specialist training strategy to enhance expert accuracy on dataset subsets.
Extensive experiments on the CIFAR-100 dataset using a real-world edge and near-edge testbed demonstrate the superiority of our framework. 
Specifically, the proposed training strategy improves expert specialization accuracy by 4.12\% on target subsets and enhances overall accuracy by 2.76\% over static experts. 
Moreover, our method reduces latency by up to 45\% compared to edge execution, and energy consumption by up to 46\% compared to just near-edge offload.
\end{abstract}

\section{Introduction}\label{sec:introduction}

The Internet of Things (IoT) has achieved ubiquitous success in various domains, with Deep Neural Networks (DNNs) serving as a pivotal mechanism for intelligent applications~\cite{chen2024end,das2024machine,wei2025cooperative}.
With the emergence of cognitive cities, there is an intensifying demand for real-time processing of data sourced from distributed sensors.
This necessitates the adoption of edge computing to mitigate the substantial communication bandwidth requirements and latency costs inherent in remote cloud processing.
However, the rapid growth in DNN parameters, computational demand, and structural complexity increasingly hinders the deployment of cutting-edge models on traditional resource-constrained edge platforms, such as the Raspberry Pi.

\begin{figure}[t] 
\centering 
\includegraphics[width=1\linewidth]{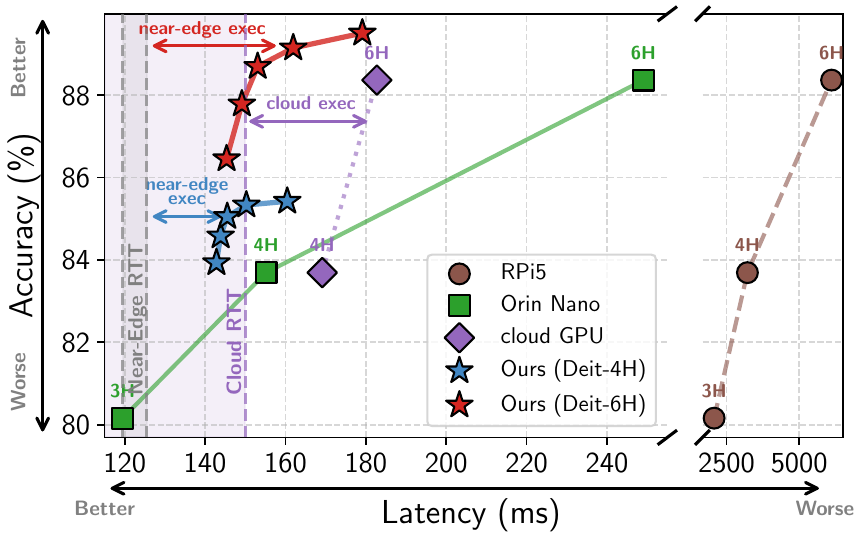}
\caption{Latency-accuracy trade-off comparison for varying model sizes (DeiT-3/4/6H) on edge (RPi5 or Orin Nano), cloud (V100), and our framework. Data points represent these models in increasing order of complexity and accuracy.}
\label{fig:motivation_tradeoff} 
\end{figure}

Recent advances in edge accelerators, such as Nvidia Jetson Orin Nano~\cite{nvidia_jetson_orinnano}, AGX Orin~\cite{nvidia_jetson_agxorin}, DGX Spark~\cite{nvidia_dgx_spark}, FPGAs~\cite{AMD_EvalBoards_2025}, or in-memory computing devices~\cite{sebastian2020memory} illustrate a continuing trend of intermediate-tier hardware situated between the edge and cloud.
These advanced devices substantially improve on-device compute capability and energy efficiency, enabling deployment of more advanced DNN models, such as Vision Transformers (ViTs)~\cite{dosovitskiy2021image,liu2021swin} near the edge.
Table~\ref{tab:execution_latency} compares the execution latency of ViT models of various sizes on different device classes, showing that modern advanced edge platforms narrow the gap with datacenter class GPUs for inference.
However, due to their significantly higher cost, we argue that such accelerators are better utilized as shared offload targets in a distributed inference setting.
This approach significantly reduces communication latency compared to remote cloud offloading, while maintaining inference speeds far superior to traditional edge devices, effectively amortizing the hardware cost.

\begin{table}[tbp]
    \centering
    \caption{Execution latency (ms) on various devices with Batch=10.}
    \footnotesize
    \label{tab:execution_latency}
    \renewcommand{\arraystretch}{1.2} 
    \begin{tabular}{@{}lcccc@{}}
        \toprule
        \multirow{2}{*}{\textbf{Variant}} & \textbf{RPi5} & \textbf{Orin Nano} & \textbf{AGX Orin} & \textbf{V100} \\
         & \textit{\$90} & \textit{\$249} & \textit{\$2k} & \textit{$>$\$10k} \\
        \midrule
        DeiT-3H     &  691.1  & 45.5  & 17.9 & 11.0  \\
        DeiT-4H     &  1073.1  & 57.1  & 20.0 & 11.2  \\  
        DeiT-6H     &  2037.7  & 88.7  & 27.6 & 15.1 \\
        DeiT-Base   &  6751.9 & 223.0 & 57.7 & 34.0 \\
        \midrule
        \textbf{Speedup}   &  1$\times$  & 15--30$\times$  & 38--117$\times$  & 63--199$\times$  \\
        \bottomrule
    \end{tabular}
    \scriptsize RPi5 latencies scaled from Batch=1 due to memory constraints. Model variants in Section~\ref{sec:experimental_setting}.
\end{table}

ViTs leverage self-attention mechanisms to capture global context, offering superior accuracy at the cost of higher computational complexity and more model parameters than traditional Convolutional Neural Networks (CNNs).
Despite the enhanced capabilities of modern edge devices, complex ViTs continue to be challenging to deploy at the edge.
These hardware and model trends motivate an ``edge to near-edge'' collaborative inference paradigm, where capable edge devices deploy simpler ViTs while leveraging more accurate complex ViTs on more capable near-edge accelerators. 
By jointly exploiting data proximity at the edge and the higher computational capacity of near-edge accelerators, collaborative inference further improves overall inference efficiency.
However, designing such an inference scheme is non-trivial, as it requires not only exploiting the varying compute capacities of heterogeneous devices but also ensuring the co-inference mechanism aligns with the unique architectural characteristics of ViTs.

Collaborative inference paradigms such as split inference and early exits have proven effective for CNNs.
However, they are ill-suited for ViTs due to inherent architectural differences.
Split inference partitions a DNN into segments for separate execution, reducing communication overhead via natural bottlenecks or compression.
In contrast, ViTs lack natural bottlenecks due to their constant columnar structure~\cite{im2024attention,jiang2025janus}, while their global self-attention induces complex dependencies that render conventional autoencoders ineffective for compression.
Alternatively, early exits terminate easy samples locally while offloading uncertain samples. However, shallow ViT layers lack sufficiently expressive representations required for accurate decisions~\cite{xu2023lgvit}.

Mixture-of-Experts (MoE)~\cite{riquelme2021scaling,fan2022m3vit,chen2023adamv,han2025vimoe} offers a promising paradigm for scaling the capacity of a single ViT while preserving comparable inference efficiency.
This paradigm replaces Feed-Forward Network (FFN) blocks with multiple parallel expert FFNs, and employs a learnable router to activate only a subset of experts for each token during inference, with outputs aggregated to yield the final results.
However, MoE models suffer from two primary challenges: training instability arising from the coupled optimization of routers and experts, which manifests as expert under-specialization and routing fluctuations~\cite{zoph2022st,dai2022stablemoe}, and router collapse, where the router consistently selects only a few experts, resulting in severe load imbalance during inference~\cite{fedus2022switch}.
Furthermore, the centralized nature of MoE models hinders their application in distributed environments~\cite{feng2025moee}.

Inspired by the MoE paradigm, we propose a system-informed approach that exploits the computational capabilities of modern near-edge hardware to improve efficiency and inference accuracy.
To this end, we propose a novel collaborative inference framework for ViTs, which orchestrates a lightweight ViT at the edge and multiple medium-sized expert ViTs on a near-edge accelerator, unlike traditional distribution of a single monolithic DNN.
Crucially, the framework decouples the router and experts, bypassing the instability of coupled optimization inherent in standard MoEs.
We introduce a novel routing mechanism that repurposes the edge ViT's Top-$\mathit{k}$ predictions as a zero-overhead router, ensuring robust expert selection that seamlessly aligns with the edge and near-edge topology.
During inference, high-confidence samples are resolved locally at the edge, while uncertain ones are routed to the specific expert determined by this lightweight router.
We additionally propose a progressive specialist training strategy to enforce expert specialization while preserving overall accuracy.

Fig.~\ref{fig:motivation_tradeoff} illustrates the latency-accuracy trade-off across varying ViT model sizes on edge, cloud, and using our collaborative framework.
Execution on a traditional edge device, such as the Raspberry Pi 5, results in prohibitive latency (over 2000ms), while moving to advanced edge hardware like the Nvidia Orin Nano significantly reduces inference time, though latency remains substantial for large models such as DeiT-6H.
When offloading from the edge, we must consider round trip latency (RTT) for the near-edge and cloud (which we take from~\cite{coll2022end}).
While the cloud GPU (V100 in this case) offers low execution time, the 150ms round trip is significant.
Our framework occupies a strategic ``sweet spot'', adding a modest near-edge execution overhead and low near-edge RTT to the local DeiT-3H processing, thereby achieving higher accuracy and lower latency than local execution or cloud offloading. The code for this paper will be open-sourced upon acceptance.

Our contributions in this paper are:
\begin{smartlist}
    \item Propose a novel collaborative inference framework for ViTs with a general formulation, leveraging a lightweight edge model and multiple near-edge experts to enhance inference efficiency and accuracy.
    \item Design a robust routing mechanism that exploits the Top-$\mathit{k}$ predictions of the edge model to determine expert activation without overhead.
    \item Propose a progressive specialist training strategy, which explicitly enforces expert specialization while preserving general accuracy.
    \item Conduct extensive experiments on CIFAR-100 dataset using a real-world edge and near-edge testbed, with results validating our training strategy for experts and showing overall accuracy gains over generalist co-inference, while reducing latency by up to 45\% and energy consumption by 46\% compared to single-device baselines.
\end{smartlist}

\section{Related Work}\label{sec:related_work}

Monolithic model distribution such as split inference~\cite{kang2017neurosurgeon,kakolyris2023road,liu2024split,ghosh2024partnner,alshams2024donna,shi2025joint,jung2025split,Chen2025CollabTrans,jiang2025janus} and early exits~\cite{li2019edge,laskaridis2020spinn,nimi2023factionformer,chen2025ceed} takes a single model and attempts to distribute it across constrained and more capable resources, but this fails to scale to more complex models, due to the lack of natural bottlenecks or compressed intermediate representations.
A variety of binary edge/cloud selection mechanisms have been explored.
\cite{li2021appealnet,cao2023edge,cao2024edge,hu2025laecips} utilize difficulty discriminators to offload complex inputs to the cloud while processing easy samples locally.
\cite{li2021appealnet} introduced a two-head CNN that simultaneously performs inference and evaluates input difficulty.
\cite{hu2025laecips} designed a ResNet-based discriminator for cloud offloading.
\cite{cao2023edge} proposed discriminators based on object statistics, while \cite{cao2024edge} introduced bandwidth adaptation.
However, these approaches require additional discriminators~\cite{li2021appealnet,hu2025laecips}, or are designed for specific tasks~\cite{cao2023edge,cao2024edge}. 
Furthermore, the high computational cost of these remote models prohibits near-edge deployment.
To further enhance local efficacy, \cite{yang2023edgefm,zhao2025c} proposed customizing lightweight edge models.
\cite{yang2023edgefm} selected edge models based on resource constraints, while \cite{zhao2025c} constructed the edge model adapted to the local data distribution.
In contrast, our framework considers a more general scenario, independent of edge resource constraints or biased data distributions.
Moving beyond edge-or-cloud offloading, \cite{eshratifar2020runtime,ma2024ditmos} employ trained multiplexers to route inputs to specialist models.
However, akin to discriminators, these methods rely on external multiplexers, which introduce additional inference overhead.

Collaborative inference has also gained traction in the realm of Large Language Models (LLMs).
Techniques such as task decomposition~\cite{narayan2025cost}, speculative decoding~\cite{ning2025dssd,xie2025novel} and skeleton completion~\cite{zhang2024cogenesis,hao2024hybrid} coordinate lightweight models to draft responses, subsequently refined by larger remote models.
These advancements underscore the growing consensus on the necessity of collaborative paradigms to address the deployment challenges posed by the escalating scale of modern models.

To reduce communication, prior works also explored selectively transmitting critical tokens via patch pruning~\cite{busto2024collaborative}, image reconstruction~\cite{liu2023efficient,liu2024adaptive}, or attention-guided image offload~\cite{im2024attention,jiang2026hyperion}. 
Since these methods primarily focus on data transmission efficiency, they are orthogonal to our proposed approach and can be potentially integrated.

Parallel to edge-cloud collaboration, some works explored distributed inference by coordinating multiple models within the cloud~\cite{hinton2015distilling} or across edge devices~\cite{xu2023devit,wang2023mdp,liu2025efficient,xu2025coformer,wen2025easyvit,feng2025moee} .
\cite{hinton2015distilling} employed a generalist model to identify a relevant specialist model for training parallelism.
\cite{xu2023devit,wang2023mdp,liu2025efficient,xu2025coformer} utilized knowledge distillation to train sub-models, incorporating class-based pruning~\cite{liu2025efficient} or boosting distillation~\cite{wang2023mdp,xu2025coformer}.
Other works optimized collaboration among identical models via token dropping~\cite{wen2025easyvit} or among heterogeneous pre-trained models via model selection~\cite{feng2025moee}.
However, these works typically require the concurrent execution and aggregation of multiple models.
In contrast, our framework routes inputs only to a single relevant expert, thereby maximizing efficiency without engaging an ensemble.

Drawing on Curriculum Learning (CL)~\cite{bengio2009curriculum,li2023curriculum}, we adopt the broader formulation of dynamic reweighting over training steps~\cite{wang2021survey,soviany2022curriculum}. 
However, unlike prior methods that reweight samples to handle noise or class imbalance~\cite{ren2018learning,jiang2018mentornet}, our progressive specialist training utilizes this mechanism to explicitly enforce domain specialization, enabling experts to transition from early-stage global generalization to late-stage specialist proficiency.

\section{Motivation}\label{sec:motivation}

\begin{figure}[t] 
\centering 
\includegraphics[width=0.8\linewidth]{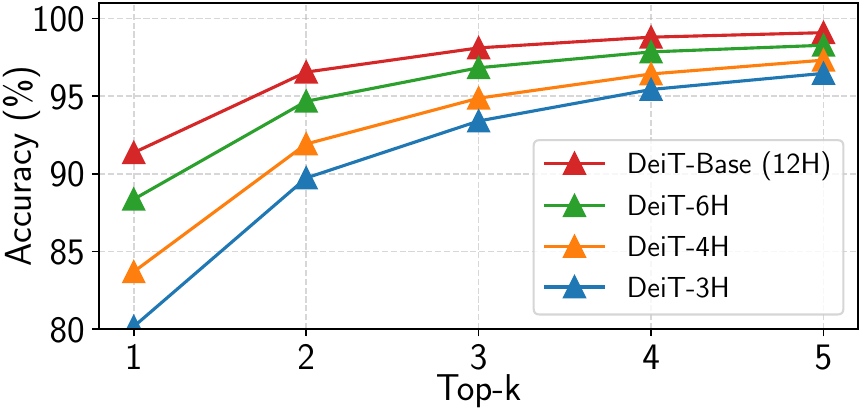}
\caption{Top-$\mathit{k}$ accuracy of ViT models of various sizes on CIFAR-100 dataset.} 
\label{fig:motivation} 
\end{figure}

We empirically evaluate ViT models of various sizes on the CIFAR-100 dataset, as depicted in Fig.~\ref{fig:motivation}. 
We distill smaller models from the original 12-head DeiT-Base.
Our experiments reveal a critical observation that while lightweight ViTs struggle to achieve high Top-1 accuracy, they exhibit significantly higher recall in their Top-$\mathit{k}$ predictions. 
For instance, the Top-1 accuracy of DeiT-3H is only 80.2\%, yet for Top-2 this reaches 89.7\%, and 93.4\% for Top-3, exceeding the Top-1 accuracy (91.4\%) of the original DeiT-Base.

This insight motivates our proposed collaborative inference framework, which aims to convert the high potential recall of lightweight ViTs at the edge into high final accuracy by refining predictions within this limited Top-$\mathit{k}$ candidate set through offloading of uncertain samples.
Crucially, since discrimination is restricted to a small subset of classes rather than the entire label space, a medium-sized expert can distinguish among these Top-$\mathit{k}$ classes more accurately than a large model across the whole dataset.
This achieves superior performance by concentrating computational resources of a near-edge accelerator on discrimination among a smaller set of candidate labels, thereby maximizing accuracy gain through targeted refinement without the overhead of a large generalist model.

\section{Proposed Method}\label{sec:proposed_method}

\subsection{Overview}\label{sec:overview}

\begin{figure}[t] 
\centering 
\includegraphics[width=1\linewidth]{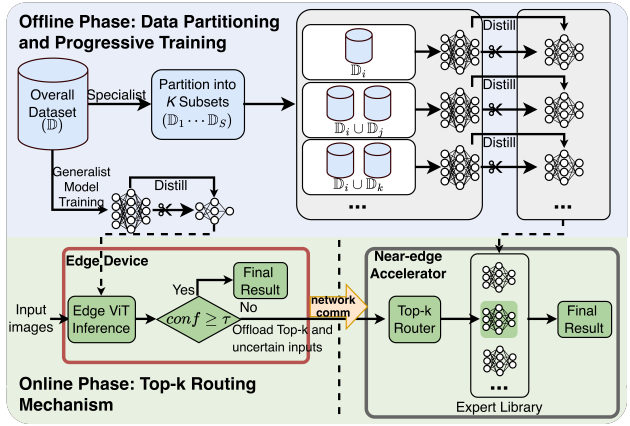}
\caption{Overview of the proposed framework.} 
\label{fig:overview} 
\end{figure}

An overview of our proposed collaborative inference framework is illustrated in Fig.~\ref{fig:overview}.
For clarity, we denote sets using blackboard-bold letters (e.g., $\mathbb{D}$ and $\mathbb{S}$),  scalars using italic letters (e.g., $\mathit{S}$ and $\mathit{k}$), and functions using calligraphic letters (e.g., $\mathcal{T}$ and $\mathcal{S}$).
The system configuration we consider in our work comprises an edge device and a near-edge accelerator.
One lightweight generalist ViT model is deployed on the edge device, while a library of expert ViT models is deployed on the near-edge accelerator.
During the offline phase, we partition the dataset into $\mathit{S}$ non-overlapping subsets.
To achieve high-precision refinement specifically for the classes in the edge model's Top-$\mathit{k}$ predictions, we propose a progressive specialist training strategy to train specialist ViT experts on flexible combinations of these subsets.
These combinations can range from a single subset to composite subsets, ensuring that a corresponding expert is available for any given Top-$\mathit{k}$ candidate set.
Simultaneously, a lightweight generalist model is trained on the entire dataset $\mathbb{D}$ to serve as the edge model.

During the online phase, the edge model first processes the inputs locally.
Using the maximum softmax probability as the confidence score, predictions exceeding a pre-defined confidence threshold $\tau$ are accepted and finalized locally, while uncertain samples are offloaded alongside their Top-$\mathit{k}$ indices to the near-edge accelerator.
The Top-$\mathit{k}$ routing mechanism maps the inputs directly to the corresponding specialist expert, which then performs the final inference to refine the classification.

\subsection{Problem Formulation}\label{sec:problem_formulation}

Following the workflow described above, we formally define the collaborative inference process.
Given an input batch of size $\mathit{B}$, let $\alpha(\tau) \in [0, 1]$ denote the offload proportion, representing the fraction of samples with confidence scores below threshold $\tau$.
Consequently, the number of samples offloaded to the near-edge accelerator is formulated as $\mathit{B}_{\mathit{off}} = \mathit{B} \cdot \alpha(\tau)$.

The overall execution latency $\mathit{T}_{\mathit{exec}}$ consists of the execution of the edge model and near-edge experts.
We profile the latency of the edge and near-edge accelerators as a function of batch size $\mathit{b}$, denoted as $\mathcal{T}_\mathit{E}(\mathit{b})$ and $\mathcal{T}_\mathit{N}(\mathit{b})$ respectively.
Thus, the edge execution latency is $\mathit{T}_{\mathit{E}} = \mathcal{T}_\mathit{E}(\mathit{B})$, while the near-edge execution latency is $\mathit{T}_{\mathit{N}} = \mathcal{T}_\mathit{N}(\mathit{B}_{\mathit{off}})$.
The overall execution latency is the sum of edge and near-edge execution latencies $\mathit{T}_{\mathit{exec}} = \mathit{T}_{\mathit{E}} +  \mathit{T}_{\mathit{N}}$, which captures the reduced set of offloaded samples.
We apply the same profiling approach to determine the overall execution energy as the sum of edge and near-edge execution energies $\mathit{E}_{\mathit{exec}} = \mathit{E}_{\mathit{E}} + \mathit{E}_{\mathit{N}}$.

Edge-cloud systems typically model communication latency $\mathit{T}_{\mathit{comm}}$ linearly as the ratio of data volume to network bandwidth~\cite{hu2019dynamic, liang2023dnn}.
However, our experiments reveal that this theoretical assumption is oversimplified for resource-constrained edge and near-edge platforms, where system overheads such as network interface packet processing are non-negligible.
Thus, instead of relying on an inaccurate theoretical model, we use the offload proportion $\alpha(\tau)$ to directly quantify the reduced data transmission volume.

\subsection{Progressive Specialist Training Strategy}~\label{sec:progressive_specialist_training}

\smartparagraph{Data Partitioning and Recomposition}~\label{sec:data_partitioning}
We partition the overall dataset $\mathbb{D}$ into $\mathit{S}$ non-overlapping subsets $\mathbb{D} = \{\mathbb{D}_1, \mathbb{D}_2, \dots, \mathbb{D}_\mathit{S}\}$ based on superclasses as in~\cite{zhao2025c}, each corresponding to a set of classes.
Let $\mathbb{S} = \{1, \dots, \mathit{S}\}$ denote the set of indices for these partitions.
Formally, let $\mathbb{I} \subseteq \mathbb{S}$ be a non-empty subset of indices representing a specific combination.
To accommodate the Top-$\mathit{k}$ routing mechanism, the cardinality $|\mathbb{I}|$ is restricted to the range $1 \le |\mathbb{I}| \le \mathit{k}$.
For each valid combination $\mathbb{I}$, we construct a composite training set $\mathbb{D}_{\mathbb{I}} = \bigcup_{\mathit{j} \in \mathbb{I}} \mathbb{D}_\mathit{j}$.
Consequently, the total number of specialist experts required is $\sum_{i=1}^{\mathit{k}} \binom{\mathit{S}}{i}$, ensuring coverage for all potential routing outcomes.

\smartparagraph{Dynamic Weighted Distillation}~\label{sec:dynamic_weighted_distillation}
Our target is to train experts $\mathbf{m}_{\mathbb{I}}$  to maximize accuracy on their designated subset $\mathbb{D}_{\mathbb{I}}$, while preserving generalization capability across the entire dataset $\mathbb{D}$.
To achieve this balance, we propose a progressive weighted specialist distillation strategy that dynamically adjusts the importance of target subsets $\mathbb{D}_{\mathbb{I}}$ during training on the entire dataset $\mathbb{D}$.
As in~\cite{xu2023devit}, for each expert model $\mathbf{m}_{\mathbb{I}}$, we employ a pre-trained large ViT model $\mathbf{M}_{\mathbb{I}}$ as the teacher.

Let $\mathit{W}$ denote the pre-defined maximum weight for the target subset.
We define a time-dependent scaling factor $\omega(\mathit{t})$ that increases linearly over the current epoch $\mathit{t}$ and total epochs $\mathit{T}$:
\begin{equation}
    \omega(\mathit{t}) = 1 + \frac{\mathit{t}}{\mathit{T}} \cdot (\mathit{W} - 1)
\end{equation}

By dynamically adjusting the importance of samples during distillation, this scaling factor ensures the model captures global features from the entire dataset in the early stages, while progressively shifting its focus toward the target domain to enforce specialization in later epochs.
Specifically, for each training sample $(x_i, y_i)$ sampled from the full dataset $\mathbb{D}$, we assign a dynamic weight $\mathit{w}_i^{(\mathit{t})}$ based on its membership in the expert target domain $\mathbb{D}_{\mathbb{I}}$:
\begin{equation}
    \mathit{w}_i^{(\mathit{t})} =\begin{cases}\omega(\mathit{t}), & \text{if } x_i \in \mathbb{D}_{\mathbb{I}} \\ 1, & \text{otherwise}\end{cases}
\end{equation}

For distillation loss, we utilize the hard decision of the teacher as the true label, as in~\cite{touvron2021training,xu2023devit}, 
so that the student learns from both the ground truth $y_i$ and the teacher's hard prediction $y^\mathbf{M}_i = \text{argmax}(\mathbf{M}_{\mathbb{I}}(x_i))$.
The total weighted loss at epoch $t$ is formulated as:
\begin{equation} 
\begin{split}
    \mathcal{L}^{(t)} = \sum_{(x_i, y_i) \in \mathbb{D}} w_i^{(t)} \cdot \bigg[ & \frac{1}{2} \mathcal{L}_{\text{CE}}(\mathbf{m}_{\mathbb{I}}(x_i), y_i) \\
    & + \frac{1}{2} \mathcal{L}_{\text{CE}}(\mathbf{m}_{\mathbb{I}}(x_i), y^\mathbf{M}_i) \bigg] 
\end{split}
\end{equation}
where $\mathcal{L}_{\text{CE}}$ denotes the standard Cross-Entropy loss.

We further clarify three aspects of our training regime.
First, to ensure high-quality supervision, the teacher models $\mathbf{M}_{\mathbb{I}}$ are pre-trained using this identical progressive specialist training strategy on their respective data subsets.
Second, prior to the distillation mentioned above, we initialize the student experts $\mathbf{m}_{\mathbb{I}}$ by pruning the teacher models $\mathbf{M}_{\mathbb{I}}$ based on the loss-based importance metric from~\cite{voita2019analyzing,hou2020dynabert,xu2023devit}.
Finally, we obtain the lightweight generalist edge model $\mathbf{m}_{\mathbb{D}}$ by pruning and distilling a large generalist ViT $\mathbf{M}_{\mathbb{D}}$ on the entire dataset $\mathbb{D}$, ensuring it remains more compact than the expert models.

\subsection{Top-$\mathit{k}$ Routing Mechanism}~\label{sec:top_k_routing_mechanism}

The proposed routing mechanism repurposes the edge model's Top-$\mathit{k}$ predictions as a zero-overhead signal for near-edge expert selection, as shown in Algorithm~\ref{alg:topk_routing}.
This collaborative inference process consists of three key steps.

\begin{algorithm}[t]
    \caption{Top-$\mathit{k}$ Routing Mechanism}
    \label{alg:topk_routing}
    \begin{algorithmic}[1]
        \Require Input $x$, Edge model $\mathbf{m}_{\mathbb{D}}$, Expert library $\mathbb{M}_{\text{exp}}$, Threshold $\tau$, Subset Mapping $\mathcal{S}(\cdot)$, Top-$\mathit{k}$ parameter $\mathit{k}$
        \Ensure Final predictions $\hat{y}$
    
    \State Probs $P \leftarrow \text{Softmax}(\mathbf{m}_{\mathbb{D}}(x))$ \label{line:edge_inference}
    \State $\mathit{conf} \leftarrow \max(P)$\label{line:confidence_compute}
    
    \If{$\mathit{conf} \geq \tau$}\label{line:high_confidence_start}
        \State $\hat{y} \leftarrow \text{argmax}(P)$\label{line:high_confidence_end}
    \Else
        \State $\mathbb{C}_{\text{top-}\mathit{k}} \leftarrow \text{Top-}\mathit{k}\text{ Indices}(P)$ \label{line:topkindices}
        
        \State $\mathbb{I} \leftarrow \emptyset$
        \For{$c \in \mathbb{C}_{\text{top-}\mathit{k}}$} \label{line:routing_start}
            \State $s \leftarrow \mathcal{S}(c)$ 
            \State $\mathbb{I} \leftarrow \mathbb{I} \cup \{s\}$ 
        \EndFor \label{line:routing_end}
        
        \State Select expert $\mathbf{m}_{\mathbb{I}} \in \mathbb{M}_{\text{exp}}$ \label{line:expert_selection}
        \State $\hat{y} \leftarrow \text{argmax}(\mathbf{m}_{\mathbb{I}}(x))$ \label{line:expert_inference}
    \EndIf
    \State \Return $\hat{y}$
    \end{algorithmic}
\end{algorithm}

\smartparagraph{Edge Inference and Confidence Check (line~\ref{line:edge_inference}~to~\ref{line:high_confidence_end}).}
For an input sample $x$, the edge model $\mathbf{m}_{\mathbb{D}}$ first performs local inference, determining confidence as the maximum softmax probability.
If the confidence is greater than or equal to a pre-defined threshold $\tau$, the local prediction is accepted as the final result.

\smartparagraph{Routing Signal Generation (line~\ref{line:topkindices}~to~\ref{line:routing_end}).}
If the sample is deemed uncertain ($\mathit{conf} < \tau$), we utilize the edge model's Top-$\mathit{k}$ predicted class indices, denoted as the set $\mathbb{C}_{\text{top-}\mathit{k}}$, as a routing signal.
We map each class index $c \in \mathbb{C}_{\text{top-}\mathit{k}}$ to its corresponding data partition ID using the mapping function $\mathcal{S}(\cdot)$.
The target domain $\mathbb{I}$ is then constructed as the union of these unique partition indices.
For instance, assuming $\mathit{k}=2$ and the predicted classes belong to partitions $1$ and $3$ respectively, the system derives the target domain $\mathbb{I}=\{1, 3\}$, thereby selecting the expert specialized on these two subsets.
This strategy ensures that for any $\mathbb{C}_{\text{top-}\mathit{k}}$, a sample can always be routed to a specialist expert optimized for the candidate classes.

\smartparagraph{Expert Refinement (line~\ref{line:expert_selection}~to~\ref{line:expert_inference}).}
Finally, the uncertain sample is routed to the selected expert $\mathbf{m}_{\mathbb{I}}$ in the expert library $\mathbb{M}_{\text{exp}}$ to perform the final prediction.

\section{Evaluation}\label{sec:evaluation}

\subsection{Experimental Setting}\label{sec:experimental_setting}

\begin{table}[t]
    \centering
    \caption{Model Configurations and Accuracy of generalist models.}
    \label{tab:models}
    \setlength{\tabcolsep}{4pt}
    \footnotesize
    \renewcommand{\arraystretch}{1.2}
    
    \begin{tabular}{@{}lccccc c@{}}
        \toprule
        \multirow{2.5}{*}{\textbf{Variant}} & 
        \multirow{2.5}{*}{\textbf{Layers}} & 
        \multirow{2.5}{*}{\makecell{\textbf{Embed.}\\\textbf{Dim.}}} & 
        \multirow{2.5}{*}{\makecell{\textbf{MLP}\\\textbf{Size}}} & 
        \multirow{2.5}{*}{\textbf{Heads}} & 
        \multicolumn{2}{c}{\textbf{Gen Acc (\%)}} \\
        
        \cmidrule(l){6-7}
        
         & & & & & \textbf{Top-1} & \textbf{Top-2} \\
        \midrule
        
        DeiT-3H   & 12 & 192 & 768  & 3  & 80.16 & 89.73 \\        
        DeiT-4H   & 12 & 256 & 1024 & 4  & 83.69 & 91.92 \\   
        DeiT-6H   & 12 & 384 & 1536 & 6  & 88.36 & 94.67 \\
        DeiT-Base & 12 & 768 & 3072 & 12 & 91.36 & 96.55 \\
        \bottomrule
    \end{tabular}
\end{table}

We evaluate our method using DeiT models~\cite{touvron2021training} on CIFAR-100~\cite{krizhevsky2009learning}. 
We first train DeiT-Base models as the teacher models ($\mathbf{M}_{\mathbb{D}}$ and $\mathbf{M}_{\mathbb{I}}$). 
Through a unified pruning and distillation pipeline, we then derive the lightweight edge model $\mathbf{m}_{\mathbb{D}}$ (configured as DeiT-3H) and the near-edge homogeneous experts $\mathbf{m}_{\mathbb{I}}$ (configured as  DeiT-4H or DeiT-6H), alongside their generalist baselines for comparison.
Detailed architectural configurations and generalist model accuracy are listed in Table~\ref{tab:models}.

All models were implemented using PyTorch 2.0.1 and trained on Nvidia A100 SXM and V100 SXM2 GPUs.
For inference, we deploy the system on an edge and near-edge testbed, consisting of an Nvidia Jetson Orin Nano 8GB as the edge device, and an Nvidia Jetson AGX Orin 64GB as the powerful near-edge accelerator.
For experimental configuration, we adopt the Top-$\mathit{k}$ routing mechanism with $\mathit{k}=2$, across dataset partition counts $\mathit{S} \in \{4, 6, 8\}$.
System performance is evaluated under a simulated workload where the edge device processes a continuous stream of images, with profiled inference latency and power consumption of each model on each device, averaged over 200 iterations, utilizing \texttt{jetson-stats}~\cite{bonghi_jetson_stats} for power monitoring.

\subsection{Effectiveness of Progressive Specialist Training}\label{sec:eval_progressive_specialist_training}

We first validate the proposed progressive specialist training strategy by comparing our trained experts against two baselines: \textbf{generalist models} trained on the entire dataset, and \textbf{static experts} trained without the progressive schedule.
Table~\ref{tab:expert_static_comparison} reports the average accuracy of the experts compared to their corresponding same-sized generalist baselines on both their target specialist data partitions and the overall dataset.

The progressively trained experts demonstrate superior specialization capabilities compared to the generalist baseline, achieving significant accuracy gains of 3.1--4.1\% for DeiT-4H, and 1.6--2.3\% for DeiT-6H on their target sub-datasets.
While providing modest gains (0.3--0.6\%) over static experts on the target partitions, they significantly outperform them on the overall dataset by margins of 2.2--2.8\% for DeiT-4H and 1.2--1.7\% for DeiT-6H.
Furthermore, we observe a positive correlation between $\mathit{S}$ and experts specialization performance, where a larger $\mathit{S}$ consistently yields higher accuracy on the target subsets.
These results confirm that our progressive strategy effectively strikes a vital balance, enforcing deep domain specialization while mitigating the degradation of global representations observed in static experts.

\begin{table}[t]
    \centering
    \caption{Average Accuracy comparison under varying dataset partition counts ($\mathit{S}$). The Generalist Baseline accuracy (\textbf{Gen.}) is provided in parentheses for each model. We compare \textbf{Static} vs. \textbf{Progr. (Ours)} on Target sub-datasets and Overall dataset.}
    \label{tab:expert_static_comparison}
    \setlength{\tabcolsep}{3.5pt}
    \footnotesize
    
    \begin{tabular}{@{}llcccc@{}}
        \toprule
        \multirow{2}{*}{\textbf{\makecell[l]{Model\\(Gen. Acc)}}} & 
        \multirow{2}{*}{$\mathbf{\mathit{S}}$} & 
        \multicolumn{2}{c}{\textbf{Target Acc. (\%)}} & 
        \multicolumn{2}{c}{\textbf{Overall Acc. (\%)}} \\
        \cmidrule(lr){3-4} \cmidrule(lr){5-6}
         & & \textbf{Static} & \textbf{Progr. (Ours)} & \textbf{Static} & \textbf{Progr. (Ours)} \\
        \midrule
        
        \multirow{3}{*}{\makecell[l]{\textbf{DeiT-4H}\\(Gen: 83.69)}} 
         & 4 & 86.30 & \textbf{86.75} & 78.94 & 81.15 \\
         & 6 & 86.92 & \textbf{87.53} & 78.13 & 80.89 \\
         & 8 & 87.18 & \textbf{87.81} & 78.47 & 81.22 \\
        \midrule
        
        \multirow{3}{*}{\makecell[l]{\textbf{DeiT-6H}\\(Gen: 88.36)}} 
         & 4 & 89.68 & \textbf{89.96} & 85.89 & 87.39  \\
         & 6 & 89.91 & \textbf{90.26} & 86.09 & 87.31\\
         & 8 & 90.23 & \textbf{90.66} & 85.45 & 87.16 \\
        \bottomrule
    \end{tabular}
\end{table}

\subsection{Performance Analysis under Varying Confidence Thresholds}\label{sec:performance_Analysis}

\begin{table}[t]
    \centering
    \caption{Offload proportion across varying confidence thresholds $\tau$.}
    \setlength{\tabcolsep}{3.5pt}
    \footnotesize
    \label{tab:offloading_percentage}
    \renewcommand{\arraystretch}{1.2}
    \begin{tabular}{@{}lcccccc@{}}
        \toprule
        \textbf{Conf Thres. $\tau$} & 0.99 & 0.90 & 0.80 & 0.70 & 0.60 & 0.50 \\
        \midrule
        \textbf{Offload prop. $\alpha(\tau)$ (\%)} & 74.7 & 46.2 & 35.8 & 27.8 & 20.8 & 13.3 \\
        \bottomrule
    \end{tabular}
\end{table}

We next compare four inference scenarios under varying confidence thresholds $\tau$, for different sizes of the DeiT model. The offload proportions across varying $\tau$ are shown in Table~\ref{tab:offloading_percentage}.
\begin{smartenum}
    \item \textbf{Edge-Only:} Inference runs on the edge device with no offload using a generalist model (3H/4H/6H).
    \item \textbf{Near-Edge-Only:} Inference is offloaded by sending inputs to the near-edge device which uses a generalist model (4H/6H).
    \item \textbf{Generalist Co-inference:} The edge device runs inference with a small (3H) general model and offloads uncertain samples to a near-edge generalist model (4H/6H/12H).
    \item \textbf{Specialist Co-inference (Ours):} The edge device runs inference with a small (3H) general model and offloads uncertain samples to the progressively trained near-edge expert model (4H/6H).
\end{smartenum}

\begin{figure}[t]
\centering
\includegraphics[width=1\linewidth]{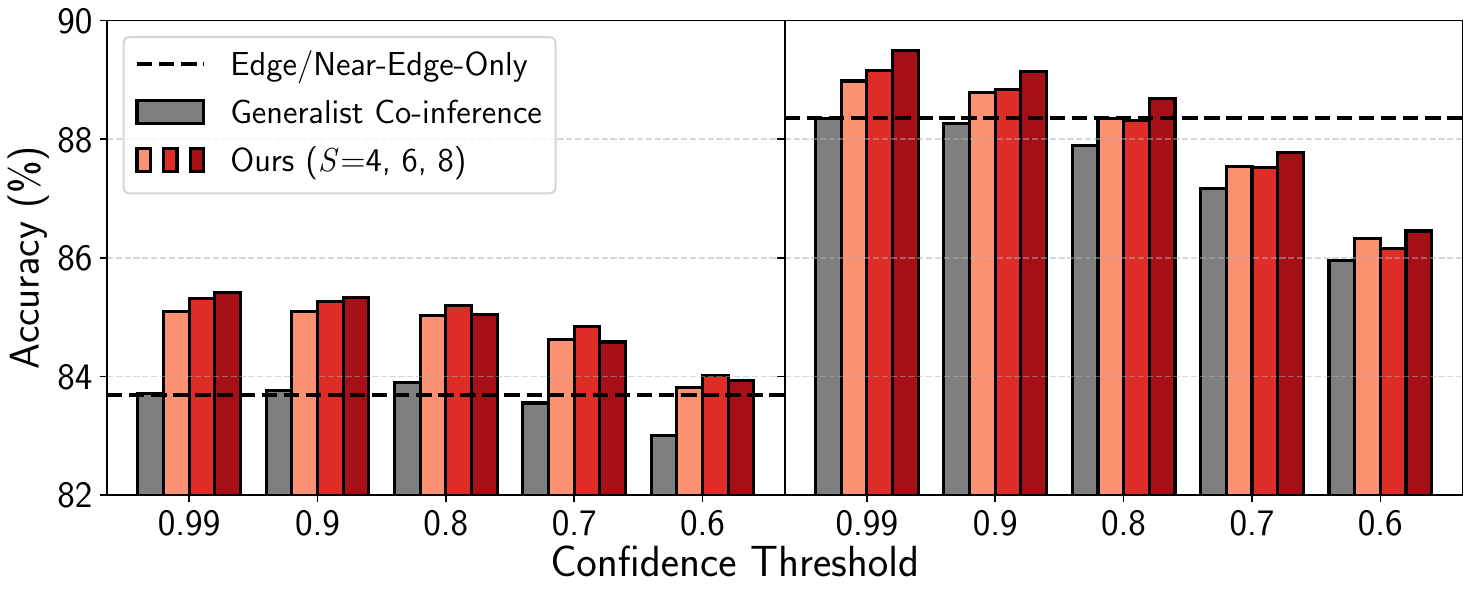} 
\caption{Accuracy comparison of Edge/Near-Edge-Only, Generalist Co-inference, and our method for DeiT-4H ($\leftarrow$) and DeiT-6H ($\rightarrow$) across varying confidence thresholds.}
\label{fig:accuracy_comparison}
\end{figure}

\smartparagraph{Accuracy Comparison.}
Fig.~\ref{fig:accuracy_comparison} illustrates the accuracy evaluation across varying confidence thresholds $\tau$.
While a lower $\tau$ reduces near-edge utilization, it introduces a marginal accuracy drop due to the lower accuracy of the edge model. 
Our method effectively mitigates this impact through expert specialization.

Our method demonstrates consistent superiority over \textit{Generalist Co-inference} with identical model scales.
As $\tau$ varies from 0.99 to 0.6, we achieve accuracy improvements of up to 1.70\% for DeiT-4H and 1.14\% for DeiT-6H across all $\mathit{S}$ settings.
This validates that the Top-$\mathit{k}$ routing mechanism selects more effective experts than using a single generalist model of the same size.
Furthermore, compared to the \textit{Edge-Only} and \textit{Near-Edge-Only}, our method offers significant gains, with improvements ranging from 0.12--1.70\% for DeiT-4H, and consistent gains of 0.33--1.14\% in high-confidence regions for DeiT-6H, despite the limited headroom left by the larger generalist model.
These results underscore the capability of our framework to exceed both standalone models and the generalist co-inference paradigm.

\begin{figure}[t]
\centering
\includegraphics[width=0.8\linewidth]{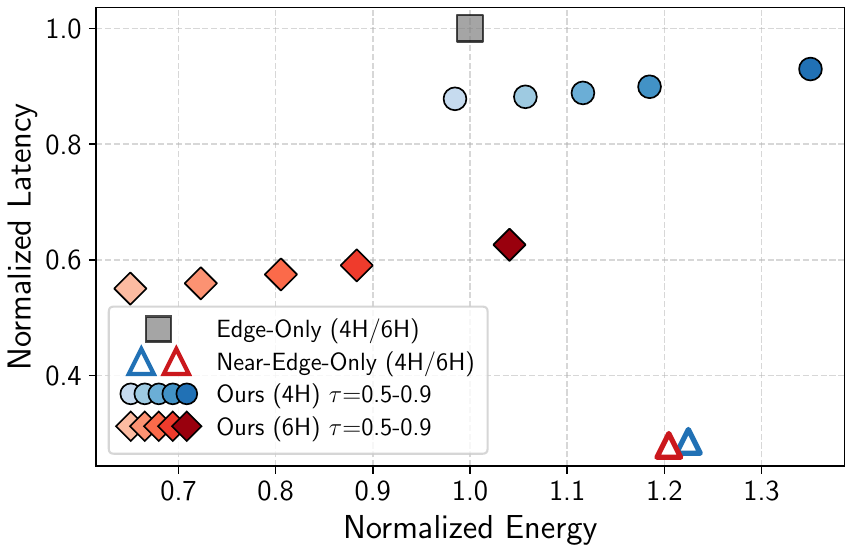} 
\caption{Normalized latency-energy comparison of Edge-Only, Near-Edge-Only, and our method for DeiT-4H and DeiT-6H across varying confidence thresholds.}
\label{fig:latency_energy_tradeoff_expert}
\end{figure}

\smartparagraph{Latency and Energy Comparison.}
We evaluate latency and energy efficiency against the \textit{Edge-Only} and \textit{Near-Edge-Only} with aligned model scales.
\textit{Generalist Co-inference} is omitted as it incurs identical overhead to our framework due to the shared underlying architecture.
Fig.~\ref{fig:latency_energy_tradeoff_expert} presents latency and energy normalized against the \textit{Edge-Only} of the corresponding generalist model across varying confidence thresholds $\tau$.
As $\tau$ decreases, fewer samples are offloaded to the near-edge device, leading to decreased latency and energy consumption.
In contrast, \textit{Edge-Only} and \textit{Near-Edge-Only} latency and energy remain constant.

Compared with \textit{Edge-Only}, our method provides a flexible trade-off between latency and energy consumption. 
For DeiT-4H, our method reduces latency by 7--12\% at the cost of a moderate energy overhead (5--35\%). 
For DeiT-6H, our method achieves a superior balance, as the prohibitive latency of local execution amplifies the benefits of offloading.
Specifically, at $\tau=0.9$, our method achieves a substantial 37\% latency reduction with a negligible 4\% energy overhead. 
When $\tau$ ranges from 0.8--0.5, our method achieves a ``win-win'' outcome, simultaneously reducing both latency by 40--45\% and energy consumption by 11--35\%.

Compared with \textit{Near-Edge-Only}, our method demonstrates significant advantages in energy efficiency, albeit with a trade-off in latency.
Specifically, our method effectively reduces energy consumption by 3--20\% for DeiT-4H, and by 14--46\% for DeiT-6H.
Regarding latency, since our method relies on local edge inference for easy samples, it incurs a fixed computational overhead. 
Hence, latency is 2--3$\times$ higher than the fully offloaded \textit{Near-Edge-Only} paradigm.
Nevertheless, our co-inference strategy is justified by the substantial reduction in communication.
As $\tau$ decreases from 0.9 to 0.5, our method drastically reduces the amount of data offloaded by 53.8--86.7\%.

\begin{figure*}[t]
\centering
\includegraphics[width=0.8\linewidth]{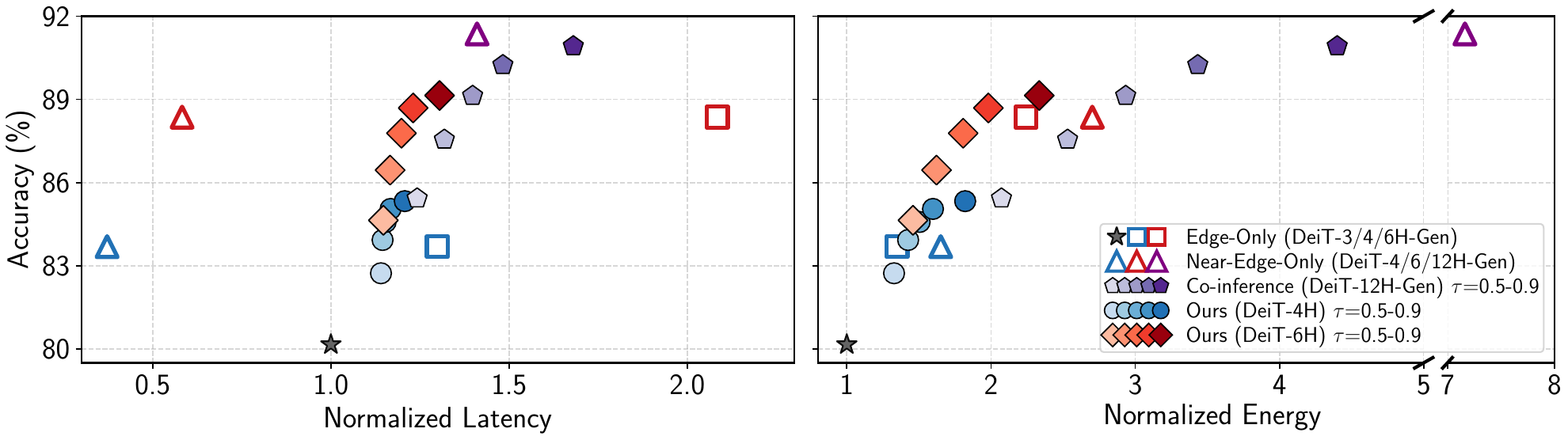} 
\caption{Latency-Accuracy ($\leftarrow$) and Energy-Accuracy ($\rightarrow$) trade-off, including Edge/Near-Edge-Only, Generalist Co-inference, and our method ($\mathit{S}=8$) across varying confidence thresholds. The latency and energy are normalized against the DeiT-3H baseline.}
\label{fig:accuracy_latency_energy_comparison_brokenaxis}
\end{figure*}

Finally, Fig.~\ref{fig:accuracy_latency_energy_comparison_brokenaxis} evaluates the global Pareto frontier by normalizing all metrics against the lightest edge model (DeiT-3H).
This holistic view confirms that our approach provides a superior trade-off between accuracy, latency, and energy across different model scales.

\begin{figure}[t]
\centering
\includegraphics[width=0.9\linewidth]{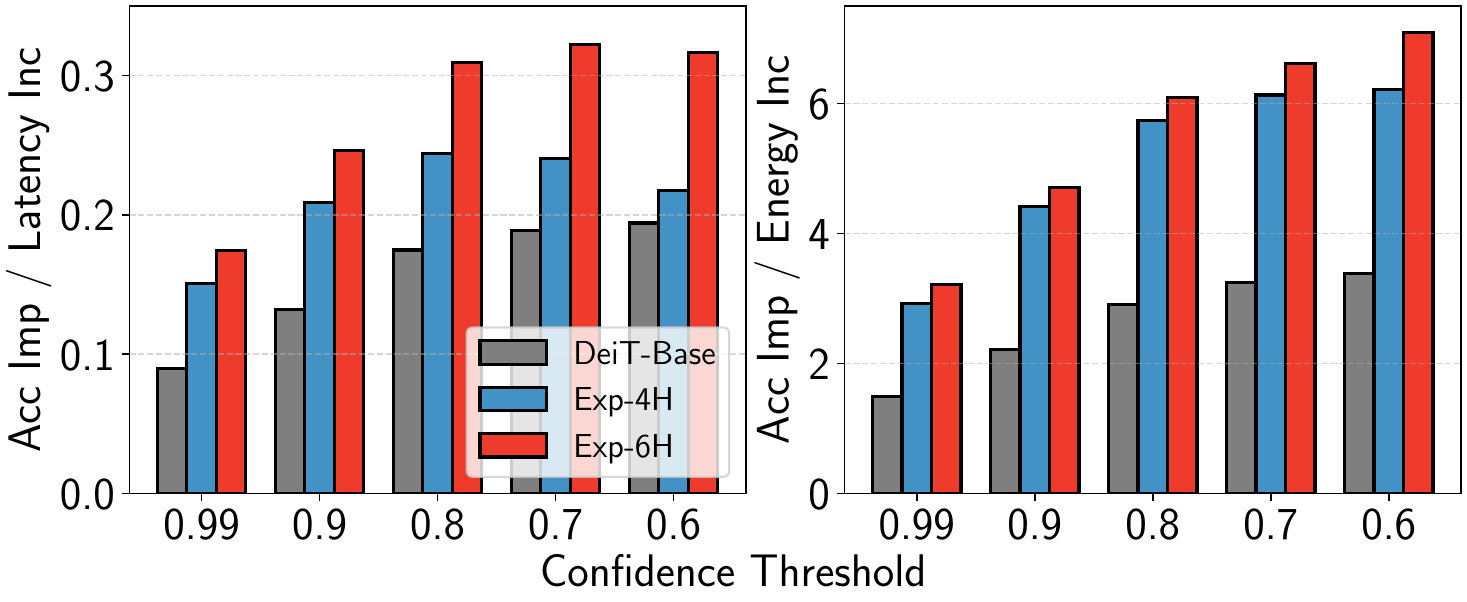} 
\caption{Experimental results ($\mathit{S}=8$) of \textit{Accuracy-to-Latency Ratio} ($\leftarrow$) and \textit{Accuracy-to-Energy Ratio} ($\rightarrow$).}
\label{fig:acc_lat_eng_ratio}
\end{figure}

\smartparagraph{Efficiency Comparison.}
To demonstrate the advantage of deploying medium-sized experts over \textit{Generalist Co-inference} with a large generalist model at the near-edge, we evaluate the accuracy improvement of this added offload over an \textit{Edge-Only} small model.
We use two metrics: \textit{Accuracy-to-Latency Ratio}  and \textit{Accuracy-to-Energy Ratio}, defined as the accuracy improvement over \textit{Edge-Only} divided by the corresponding incremental latency or energy cost.
These represent the ``return on investment'' (ROI) for offloading uncertain samples, where a higher ratio means more efficient resource utilization.
Our results are illustrated in Fig.~\ref{fig:acc_lat_eng_ratio}.

Although Generalist Co-Inference with a large model achieves higher absolute accuracy, it incurs disproportionately higher latency and energy cost.
Our method consistently outperforms Generalist Co-inference in both efficiency metrics across varying $\tau$.
These results confirm that employing specialist medium-sized experts yields a superior solution for resource-constrained near-edge accelerators compared to deploying large generalist models to support edge offloading.

\section{Discussion}\label{sec:discussion}

We conducted extensive experiments on CIFAR-100 to allow us to refine our approach and present complete results, using over 1500 GPU hours.
However, our framework is expected to yield even greater efficiency gains on more complex tasks as its effectiveness fundamentally hinges on the ``recall gap'' between the Top-$\mathit{k}$ and Top-1 accuracy.
This gap represents the ``recoverable'' portion of samples where the lightweight model correctly includes the ground truth in its candidate set but fails to select it as the final prediction.
Consequently, a larger gap indicates a higher potential for our near-edge experts to rectify the edge model's ``near-miss'' errors.
Table~\ref{tab:acc_gap_comparison} illustrates that this phenomenon is more pronounced on larger datasets, with our edge model (DeiT-3H) exhibiting a recall gap of 13.24\% between Top-3 and Top-1 accuracy on CIFAR-100, but 13.98\% on ImageNet.
This suggests that as task complexity (from 100 to 1000 classes) increases, the edge model functions increasingly as an effective semantic router, creating a larger opportunity for our specialist experts to recover accuracy that would otherwise be lost in standard monolithic inference.
Furthermore, prioritizing Top-$\mathit{k}$ accuracy over Top-1 accuracy during edge model training can further enhance the precision of our routing mechanism~\cite{petersen2022differentiable}.

The introduction of newer edge hardware like the DGX Spark presents further diversity in the cost/performance trade-off for hardware that can be deployed near the edge and serve multiple edge devices, and hence, we expect frameworks such as ours to increase in importance as the demand for ubiquitous intelligence increases, with the latency and privacy benefits of near-edge offload becoming increasingly important.
Finally, our framework is orthogonal to model compression techniques such as quantization~\cite{li2023repq} and pruning~\cite{yang2023global}, where our framework can effectively compensate for the potential accuracy degradation inherent due to aggressive compression of edge models.

\begin{table}[t]
    \centering
    \setlength{\tabcolsep}{3.5pt}
    \footnotesize
    \renewcommand{\arraystretch}{1.2}
    \caption{Top-$\mathit{k}$ accuracy and Recall gaps ($\Delta$) for DeiT-3H.}
    \label{tab:acc_gap_comparison}
    \begin{tabular}{lccccc}
        \toprule
        \multirow{2}{*}{\textbf{Dataset}} & \multicolumn{3}{c}{\textbf{Accuracy (\%)}} & \multicolumn{2}{c}{\textbf{Recall Gap  (\%)}} \\
        \cmidrule(lr){2-4} \cmidrule(lr){5-6}
        
         & \textbf{Top-1} & \textbf{Top-2} & \textbf{Top-3} & ${\Delta_{2-1}}$ & ${\Delta_{3-1}}$ \\
        \midrule
        CIFAR-100 & 80.16 & 89.73 & 93.40 & 9.57 & 13.24 \\
        ImageNet  & 74.52 & 84.62 & 88.50 & \textbf{10.10} & \textbf{13.98} \\
        \bottomrule
    \end{tabular}
\end{table}

\section{Conclusions}\label{sec:conclusion}

We have proposed a novel collaborative inference framework to address the challenges of deploying complex ViTs at the edge.
Our framework orchestrates a lightweight generalist model on the edge and multiple specialist experts on a near-edge accelerator, with two key technical innovations. 
First, the proposed routing mechanism repurposes the edge model's Top-$\mathit{k}$ predictions as a zero-overhead router, dynamically routing uncertain samples to the most relevant expert on the near-edge. 
Second, our progressive specialist training strategy effectively enforces expert specialization on data subsets while preserving overall generalization capability.

Extensive experiments on CIFAR-100 demonstrate the effectiveness of our approach. 
The training strategy improves expert specialization accuracy by 4.12\% on target subsets and enhances overall accuracy by 2.76\%.
Our framework also achieves accuracy gains of up to 1.7\% over generalist co-inference, while simultaneously reducing latency by up to 45\% and energy consumption by 46\% compared to standalone inference. 
Finally, the proposed framework yields a superior Pareto-optimal solution, validating that deploying medium-sized experts on resource-constrained near-edge accelerators is significantly more efficient for supporting edge offloading than relying on a large generalist model.

\section*{Impact Statement}

This paper introduces a framework for collaborative inference, enabling enhanced accuracy and lower latency for edge AI applications, and reducing the dependency on centralized cloud datacenters, thereby enhancing efficiency.
From an ethical and societal standpoint, this research contributes to the advancement of energy-efficient, sustainable AI, while democratizing access to advanced computer vision in remote or infrastructure-poor environments.

\bibliography{main}
\bibliographystyle{icml2026}

\newpage
\appendix
\onecolumn
\section{Appendix}

\subsection{Implementation Details}\label{app:implementation_details}

In this section, we detail the training configurations for both the generalist model and expert models. 
All experiments were conducted on CIFAR-100 using PyTorch 2.0.1. 
We initialize both the large generalist and expert teachers with the ImageNet-pretrained DeiT-Base, and derive their lightweight students through a unified pruning and distillation pipeline.
For all training phases, we employed the AdamW optimizer and a cosine learning rate scheduler with a 5-epoch linear warmup.
We disabled strong data augmentations (Mixup, Cutmix) and Label Smoothing for expert training, as they compromise the expert accuracy on the target sub-datasets.

\subsection{Training Hyperparameters}\label{app:training_params}

We summarize the hyperparameters for both the generalist and specialist models in Table~\ref{tab:unified_training_params}. 
The training involves a pre-training stage for teachers (DeiT-Base) and a distillation stage for students (DeiT-3/4/6H). 
The specialist training introduces a progressive mechanism with an empirical maximum weight parameter $\mathit{W}$, distinguishing it from the generalist distillation.

\begin{table}[htp]
    \centering
    \caption{\textbf{Unified Model Settings.} Hyperparameters for training generalist and specialist models (teachers and students), including optimization and progressive training details.}
    \label{tab:unified_training_params}
    \vspace{0.1in}
    \begin{tabular}{lcccc}
    \toprule
    & \multicolumn{2}{c}{\textbf{Generalist}} & \multicolumn{2}{c}{\textbf{Specialist}} \\
    \cmidrule(lr){2-3} \cmidrule(lr){4-5}
    \textbf{Hyperparameter} & \textbf{Teacher} & \textbf{Student} & \textbf{Teacher} & \textbf{Student} \\
    \midrule
    \textit{Model Configuration} & & & & \\
    Architecture & DeiT-Base & DeiT-(3/4/6H) & DeiT-Base & DeiT-(4/6H) \\
    \midrule
    \textit{Optimization} & & & & \\
    Batch Size & 256 & 256 & 256 & 256 \\
    Total Epochs & 200 & 200 & 200 & 200 \\
    Learning Rate (LR) & $1 \times 10^{-4}$ & $8 \times 10^{-5}$ & $1 \times 10^{-4}$ & $8 \times 10^{-5}$ \\
    Clip Grad & None & 1.0 & None & 1.0 \\
    \midrule
    \textit{Strategy \& Distillation} & & & & \\
    Distillation Type & None & Hard & None & Hard \\
    Teacher Source & N/A & Generalist & N/A & Specialist \\
    Progressive Training & N/A & N/A & True & True \\
    Max Weight ($\mathit{W}$) & -- & -- & 14.0 & 14.0 \\
    \bottomrule
    \end{tabular}
\end{table}

\subsection{Data Division Details}\label{app:data_division_details}

We provide detailed dataset partition schemes in our experiments for CIFAR-100 in Table~\ref{tab:merged_superclass_distribution}, where superclasses were randomly assigned to each partition.

\begin{table}[t] 
    \centering
    \caption{\textbf{Superclass-based CIFAR-100 Dataset Partition Schemes.} Detailed breakdown of superclass distribution across 4, 6, and 8 sub-datasets.}
    \label{tab:merged_superclass_distribution}
    
    \vspace{0.1in}
    \setlength{\tabcolsep}{4pt} 
    \footnotesize               
    \renewcommand{\arraystretch}{1.2} 

    \begin{tabular}{@{} l c >{\raggedright\arraybackslash}p{9.5cm} @{}}
    \toprule
    \textbf{Subset Index} & \textbf{Count} & \textbf{Superclasses Included} \\ 
    \midrule

    \multicolumn{3}{l}{\textit{\textbf{Partition into 4 Sub-datasets}}} \\ 
    \midrule
    1 & 5 & Fruit and vegetables, People, Large man-made outdoor things, Household electrical devices, Vehicles 2 \\ 
    2 & 5 & Household furniture, Vehicles 1, Invertebrates, Reptiles, Medium-sized mammals \\ 
    3 & 5 & Fish, Large omnivores and herbivores, Large natural outdoor scenes, Trees, Flowers \\ 
    4 & 5 & Large carnivores, Aquatic mammals, Insects, Food containers, Small mammals \\ 
    \midrule

    \multicolumn{3}{l}{\textit{\textbf{Partition into 6 Sub-datasets}}} \\ 
    \midrule
    1 & 4 & Fruit and vegetables, People, Household electrical devices, Vehicles 2 \\ 
    2 & 4 & Vehicles 1, Large man-made outdoor things, Invertebrates, Reptiles \\ 
    3 & 3 & Household furniture, Medium-sized mammals, Trees \\ 
    4 & 3 & Fish, Large omnivores and herbivores, Large natural outdoor scenes \\ 
    5 & 3 & Insects, Small mammals, Flowers \\ 
    6 & 3 & Large carnivores, Aquatic mammals, Food containers \\ 
    \midrule

    \multicolumn{3}{l}{\textit{\textbf{Partition into 8 Sub-datasets}}} \\ 
    \midrule
    1 & 3 & People, Household electrical devices, Vehicles 2 \\ 
    2 & 3 & Fruit and vegetables, Large man-made outdoor things, Invertebrates \\ 
    3 & 3 & Household furniture, Vehicles 1, Reptiles \\ 
    4 & 3 & Large natural outdoor scenes, Medium-sized mammals, Trees \\ 
    5 & 2 & Fish, Large omnivores and herbivores \\ 
    6 & 2 & Small mammals, Flowers \\ 
    7 & 2 & Insects, Large carnivores \\ 
    8 & 2 & Aquatic mammals, Food containers \\ 
    
    \bottomrule
    \end{tabular}
\end{table}

\end{document}